\documentclass[10pt,twocolumn,letterpaper]{article}

\usepackage{forloop}
\usepackage{etoolbox}

\usepackage[table]{xcolor}
\usepackage{titlesec}

\usepackage{subcaption}
\usepackage{forloop}
\usepackage{etoolbox}

\usepackage{caption}
\newcommand{\comment}[1]{}

\newcommand*\rot{\rotatebox{90}}

\def\best{\bf \cellcolor[gray]{0.85}}
\def\secbest{\cellcolor[gray]{0.92} }

\usepackage{cvpr}
\usepackage{times}
\usepackage{epsfig}
\usepackage{graphicx}
\usepackage{amsmath}
\usepackage{amssymb}
\usepackage[numbers]{natbib}

\usepackage{float}

\newcommand{\Fig}{Fig.\xspace}

\newcommand{\Sec}{Sec.\xspace}

\newcommand{\myparagraph}[1]{\vspace{.5em}\noindent\textbf{#1}}
\newcommand{\ep}[1]{\emph{(#1)}}
\newcommand{\refOurFig}{\ref{fig:refine-net}(c)\xspace}

\usepackage[pagebackref=true,breaklinks=true,letterpaper=true,colorlinks,bookmarks=false]{hyperref}

\cvprfinalcopy 

\def\cvprPaperID{***} 

\ifcvprfinal\fi

\begin{document}

\title{RefineNet: Multi-Path Refinement Networks for\\ High-Resolution Semantic Segmentation}

\author{Guosheng Lin$^{1,2}$, \; Anton Milan$^{1}$, \; Chunhua Shen$^{1,2}$, \; Ian Reid$^{1,2}$\\
$^{1}$The University of Adelaide, \; $^{2}$Australian Centre for Robotic Vision\\
{\tt\small \{guosheng.lin;anton.milan;chunhua.shen;ian.reid\}@adelaide.edu.au}
}

\maketitle

\begin{abstract}

Recently, very deep convolutional neural networks (CNNs) have shown outstanding performance in object recognition and have also been the first choice for dense classification problems such as semantic segmentation. However, repeated subsampling operations like pooling or convolution striding in deep CNNs lead to a significant decrease in the initial image resolution.  
Here, we present \emph{RefineNet}, a generic multi-path refinement network that explicitly exploits all the information available along the down-sampling process to enable high-resolution prediction using long-range residual connections. In this way, the deeper layers that capture high-level semantic features can be directly refined using fine-grained features from earlier convolutions. The individual components of RefineNet employ residual connections following the identity mapping mindset, which allows for effective end-to-end training. Further, we introduce chained residual pooling, which captures rich background context in an efficient manner. We carry out comprehensive experiments and set new state-of-the-art results on seven public datasets. In particular, we achieve an intersection-over-union score of $83.4$ on the challenging PASCAL VOC 2012 dataset, which is the best reported result to date.

\end{abstract}

\section{Introduction}
\label{sec:introduction}

Semantic segmentation is a crucial component in image understanding. The task here is to assign a unique label (or category) to every single pixel in the image, which can be considered as a dense classification problem. The related problem of so-called object parsing can usually be cast as semantic segmentation.
Recently, deep learning methods, and in particular convolutional neural networks (CNNs), \eg, VGG~\cite{simonyan2014very}, Residual Net~\cite{He:2016:ResNet}, have shown remarkable results in recognition tasks. 
However, these approaches exhibit clear limitations when it comes to dense prediction in tasks like dense depth or normal estimation~\cite{eigen2015predicting,liu2014deep,liu2015learning} and semantic segmentation~\cite{LongSD14,ChenPKMY14}. Multiple stages of spatial pooling and convolution strides reduce the final image prediction typically by a factor of 32 in each dimension, thereby losing much of the finer image structure.

\begin{figure}[t]
	\centering	
	\includegraphics[width=1\linewidth]{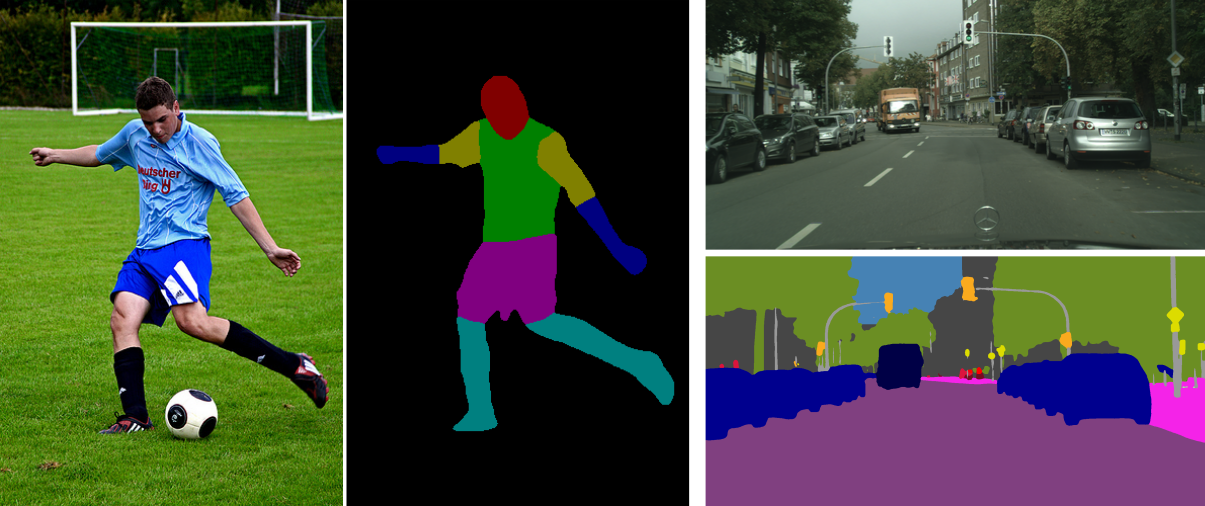}	
\caption{Example results of our method on the task of object parsing~ \ep{left} and semantic segmentation~\ep{right}.}
\label{fig:teaser}
\end{figure}

One way to address this limitation is to learn deconvolutional filters as an up-sampling operation~\cite{noh2015learning,LongSD14} to generate high-resolution feature maps. 
The deconvolution operations are not able to recover the low-level visual features which are lost after the down-sampling operation in the convolution forward stage. Therefore,
they are unable to output accurate high-resolution prediction. Low-level visual information is essential for accurate prediction on the boundaries or details.
The method DeepLab recently proposed by Chen~\etal~\cite{ChenPK0Y16} employs atrous (or dilated) convolutions to account for larger receptive fields without downscaling the image. 
DeepLab is widely applied and represents state-of-the-art performance on semantic segmentation.
This strategy, although successful, has at least two limitations. 
First, it needs to perform convolutions on a large number of detailed (high-resolution) feature maps that usually have high-dimensional features, which are computational expensive.
Moreover, a large number of high-dimensional and high-resolution feature maps also require huge GPU memory resources, especially in the training stage.
This hampers the computation of high-resolution predictions and usually limits the output size to $1/8$ of the original input.
Second, dilated convolutions introduce a coarse sub-sampling of features, which potentially leads to a loss of important details. 

Another type of methods exploits features from intermediate layers for generating high-resolution prediction, \eg, the FCN method in~\cite{LongSD14} and Hypercolumns in~\cite{hariharan2014hypercolumns}.
The intuition behind these works is that features from middle layers are expected to describe mid-level representations for object parts, while retaining spatial information. This information is though to be complementary to the features from early convolution layers which encode low-level spatial visual information like edges, corners, circles, \etc, and also complementary to high-level features from deeper layers which encode high-level semantic information, including object- or category-level evidence, but which lack strong spatial information.

We argue that features from all levels are helpful for semantic segmentation. High-level semantic features helps the category recognition of image regions, while low-level visual features help to generate sharp, detailed boundaries for high-resolution prediction.  
How to effectively exploit middle layer features remains an open question and deserves more attentions.
To this end we propose a novel network architecture which effectively exploits multi-level features for generating high-resolution predictions.
Our main \emph{contributions} are as follows:
\vspace{-2pt}
\begin{enumerate}

  \item We propose a multi-path refinement network (RefineNet) 
  which exploits features at multiple levels of abstraction for high-resolution semantic segmentation. 
  RefineNet refines low-resolution (coarse) semantic features with fine-grained low-level features in a recursive manner to generate high-resolution semantic feature maps. Our model is flexible in that it can be cascaded and modified in various ways. 

  \item 
  Our cascaded RefineNets can be effectively trained end-to-end, which is crucial for best prediction performance.
  More specifically, all components in RefineNet employ residual connections~\cite{He:2016:ResNet} with identity mappings~\cite{he2016identity},
  such that gradients can be directly propagated through short-range {\em and} long-range residual connections allowing for both effective and efficient end-to-end training.

  \item We propose a new network component we call ``chained residual pooling'' which is able to capture background context from a large image region.
	It does so by efficiently pooling features with multiple window sizes and fusing them together with residual connections and learnable weights.

  \item The proposed RefineNet achieves new state-of-the-art performance on 7 public datasets, including PASCAL VOC 2012, PASCAL-Context, NYUDv2, SUN-RGBD, Cityscapes, ADE20K, and the object parsing Person-Parts dataset. 
	In particular, we achieve an IoU score of $83.4$ on the PASCAL VOC 2012 dataset, outperforming the currently best approach DeepLab by a large margin.
  
\end{enumerate}

To facilitate future research,
we release both source code and trained models for our RefineNet.\footnote{Our source code will be available at \url{https://github.com/guosheng/refinenet}}

\subsection{Related Work}
\label{sec:related-work}

CNNs become the most successful methods for semantic segmentation in recent years.
The early methods in \cite{GirshickDDM13,BharathECCV2014} are region-proposal-based methods which classify region proposals to generate segmentation results.
Recently fully convolution network (FCNNs) based based methods \cite{LongSD14,ChenPKMY14,Dai2015arXiv} show effective feature generation and end-to-end training,
 and thus become the most popular choice for semantic segmentation.
FCNNs have also been widely applied in other dense-prediction tasks, e.g., depth estimation \cite{dcnn_nips14,eigen2015predicting,liu2014deep},
image restoration \cite{Eigen_iccv13}, image super-resolution \cite{Dong_eccv14}.
The proposed method here is also based on fully convolution-style networks.

FCNN based methods usually have the limitation of low-resolution prediction.
There are a number of proposed techniques which addressed this limitation and aim to generate high-resolution predictions.
The atrous convolution based approach DeepLab-CRF in \cite{ChenPKMY14} directly output a middle-resolution score map 
then applies the dense CRF method \cite{krahenbuhl2012efficient} to refine boundaries by leveraging color contrast information.
CRF-RNN \cite{zheng2015conditional} extends this approach by implementing recurrent layers for end-to-end learning of the dense CRF and FCNN.
Deconvolution methods~\cite{noh2015learning,BadrinarayananK15} learn deconvolution layers to up-sample the low-resolution predictions.
The depth estimation method \cite{liu2015learning} employs super-pixel pooling to output high-resolution prediction.

There are several existing methods which exploit middle layer features for segmentation. The FCN method in \cite{LongSD14} adds prediction layers to middle layers to generate prediction scores at multiple resolutions. They average the multi-resolution scores to generate the final prediction mask. Their system is trained in a stage-wise manner rather than end-to-end training. The method Hypercolumn \cite{hariharan2014hypercolumns} merges features from middle layers and learns dense classification layers. Their method employs stage-wise training instead of end-to-end training. The method Seg-Net \cite{BadrinarayananK15} and U-Net \cite{Ronneberger2015} apply skip-connections in the deconvolution architecture to exploit the features from middle layers.

Although there are a few existing work, how to effectively exploit middle layer features remains an open question.
We propose a novel network architecture, RefineNet, to address this question.
The network architecture of RefineNet is clearly different from existing methods.
RefineNet consists of a number of specially designed components which are able to refine the coarse high-level semantic features by exploiting low-level visual features.
In particular, RefineNet employs short-range and long-range residual connections with identity mappings 
which enable effective end-to-end training of the whole system, and thus help to archive good performance. 
Comprehensive empirical results clearly verify the effectiveness of our novel network architecture for exploiting middle layer features.

\section{Background}
\label{sec:background}

\begin{figure*}[t]
	\centering	
	\includegraphics[width=1\linewidth]{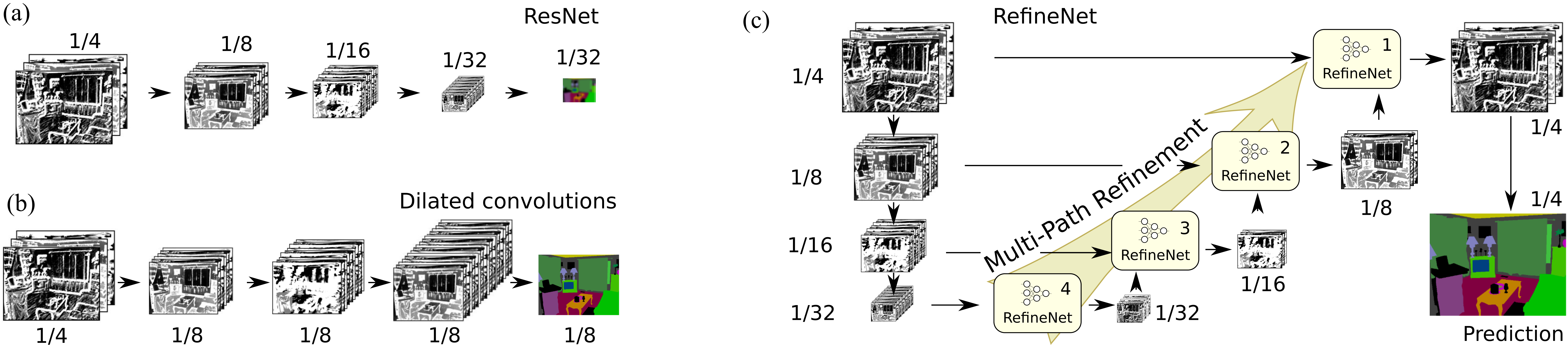}	
\caption{Comparison of fully convolutional approaches for dense classification. 
Standard multi-layer CNNs, such as ResNet \ep{a} suffer from downscaling of the feature maps, thereby losing fine structures along the way. 
Dilated convolutions \ep{b} remedy this shortcoming by introducing atrous filters, but are computationally expensive to train and quickly reach memory limits even on modern GPUs.
Our proposed architecture that we call RefineNet \ep{c} exploits various levels of detail at different stages of convolutions and fuses them to obtain a high-resolution prediction without the need to maintain large intermediate feature maps. The details of the RefineNet block are outlined in \Sec~\ref{sec:refine-net} and illustrated in Fig~\ref{fig:refine-net-detailed-joint}.}
\label{fig:refine-net}
\end{figure*}

Before presenting our approach, we first review the structure of fully convolutional networks for semantic segmentation~\cite{LongSD14} in more detail and also discuss the recent dilated convolution technique~\cite{ChenPK0Y16} which is specifically designed to generate high-resolution predictions.

Very deep CNNs
have shown outstanding performance on object recognition problems.
Specifically, the recently proposed Residual Net (ResNet)~\cite{He:2016:ResNet} has shown step-change improvements over earlier architectures, and ResNet models pre-trained for ImageNet recognition tasks are publicly available.
Because of this, in the following we adopt ResNet as our fundamental building block for semantic segmentation. Note, however, that replacing it with any other deep network is straightforward.

Since semantic segmentation can be cast as a dense classification problem, the ResNet model can be easily modified for this task.
This is achieved by replacing the single label prediction layer with a dense prediction layer that outputs the classification confidence for each class at every pixel.
This approach is illustrated in~\Fig~\ref{fig:refine-net}(a).
As can be seen, during the forward pass in ResNet, the resolution of the feature maps (layer outputs) is decreased, while the feature depth, \ie the number of feature maps per layer (or {\em channels}) is increased.
The former is caused by striding during convolutional and pooling operations.

The ResNet layers can be naturally divided into 4 blocks according to the resolution of the output feature maps, as shown in ~\Fig~\ref{fig:refine-net}(a).
Typically, the stride is set to 2, thus reducing the feature map resolution to one half when passing from one block to the next. This sequential sub-sampling has two effects: first it increases the receptive field of convolutions at deeper levels, enabling the filters to capture more global and contextual information which is essential for high quality classification; second it is necessary to keep the training efficient and tractable because each layer comprises a large number of filters and therefore produces an output which has a corresponding number of channels, thus there is a trade-off between the number of channels and resolution of the feature maps. Typically the final feature map output ends up being 32 times smaller in each spatial dimension than the original image (but with 1000s of channels). 
This low-resolution feature map loses important visual details captured by early low-level filters, resulting in a rather coarse segmentation map.
This issue is a well-known limitation of deep CNN-based segmentation methods.

An alternative approach to avoid lowering the resolution while retaining a large receptive field is to use dilated (atrous) convolution.  This method introduced in \cite{ChenPK0Y16}, has the state-of-the-art performance on semantic segmentation.  The sub-sampling operations are removed (the stride is changed from 2 to 1), and all convolution layers after the first block use dilated convolution.  Such a dilated convolution (effectively a sub-sampled convolution kernel) has the effect of increasing the receptive field size of the filters without increasing the number of weights that must be learned (see illustration in \Fig~\ref{fig:refine-net}(b)). Even so, there is a significant cost in memory, because unlike the image sub-sampling methods, one must retain very large numbers of feature maps at higher resolution.  For example, if we retain all channels in all layers to be at least 1/4 of the original image resolution, and consider a typical number of filter channels to be 1024, then we can see that the memory capacity of even high-end GPUs is quickly swamped by very deep networks.  In practice, therefore, dilation convolution methods usually have a resolution prediction of no more than 1/8 size of the original rather than 1/4, when using a deep network.

In contrast to dilated convolution methods, in this paper we propose a means to enjoy both the memory and computational benefits of deresolving, while still able to produce effective and efficient high-resolution segmentation prediction, as described in the following section.

\begin{figure*}[t]
	\centering	
	\includegraphics[width=1\linewidth]{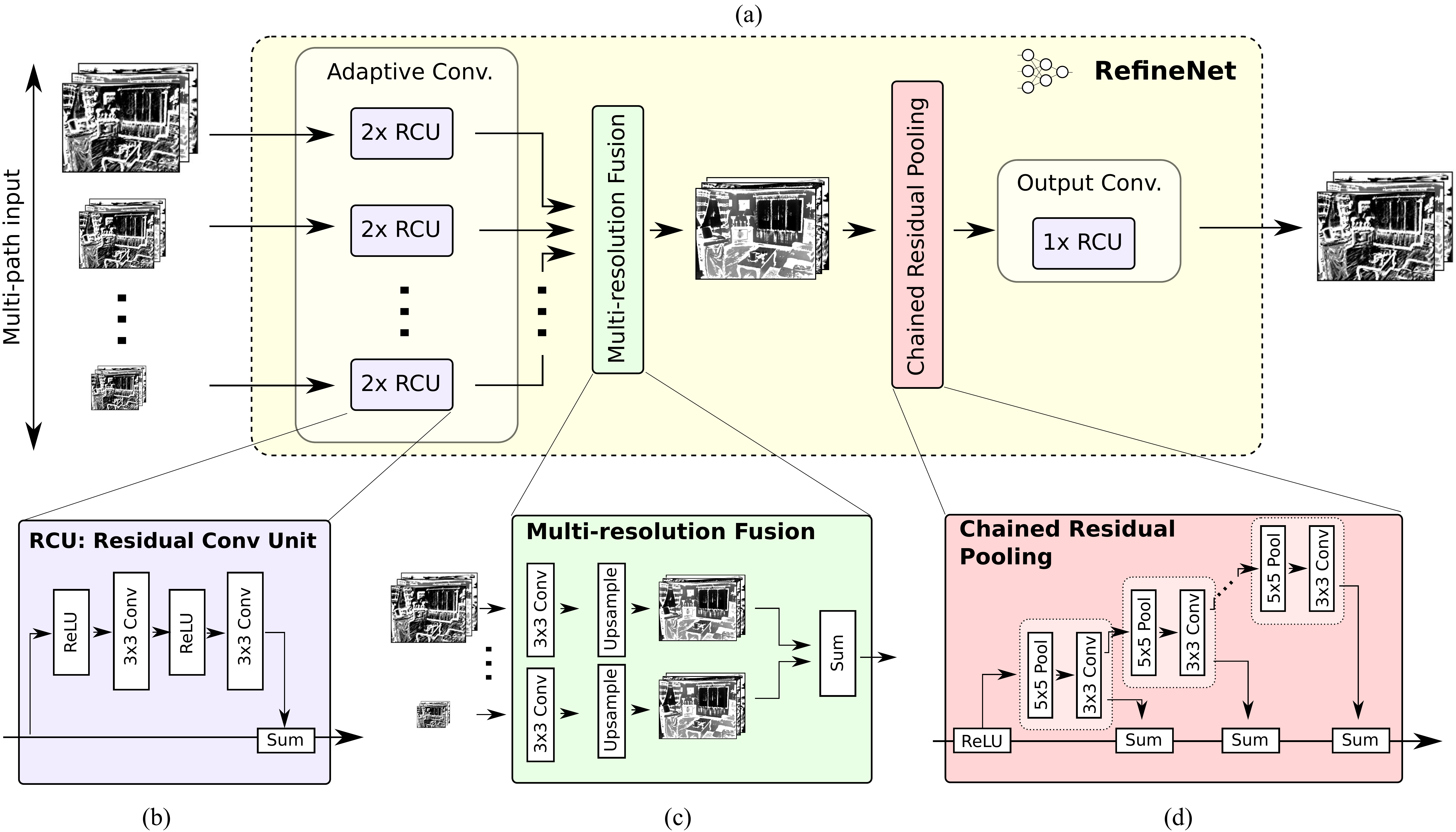}	
\caption{The individual components of our multi-path refinement network architecture RefineNet. Components in RefineNet employ residual connections with identity mappings. In this way, gradients can be directly propagated  within RefineNet via local residual connections, and also directly propagate to the input paths via long-range residual connections, 
and thus we achieve effective end-to-end training of the whole system.}
\label{fig:refine-net-detailed-joint}
\end{figure*}

\section{Proposed Method}
\label{sec:refine-net}

We propose a new framework that provides multiple paths over which information from different resolutions and via potentially long-range connections, is assimilated using a generic building block, the RefineNet. \Fig~\refOurFig shows one possible arrangement of the building blocks to achieve our goal of high resolution semantic segmentation.  We begin by describing the multi-path refinement arrangement in \Sec~\ref{sec:method-overview} followed by a detailed description of each RefineNet block in \Sec\ref{sec:refinet_details}.

\subsection{Multi-Path Refinement}
\label{sec:method-overview}

As noted previously, we aim to exploit multi-level features for high-resolution prediction with long-range residual connections. 
RefineNet provides a generic means to fuse coarse high-level semantic features with finer-grained low-level features to generate high-resolution semantic feature maps. A crucial aspect of the design ensures that the gradient can be effortlessly propagated backwards through the network all the way to early low-level layers over long-range residual connections, ensuring that the entire network can be trained end-to-end.

For our standard multi-path architecture, we divide the pre-trained ResNet (trained with ImageNet) into 4 blocks according to the resolutions of the feature maps, and employ a 4-cascaded architecture with 4 RefineNet units, each of which directly connects to the output of one ResNet block as well as to the preceding RefineNet block in the cascade. 
Note, however, that such a design is not unique. In fact, our flexible architecture allows for a simple exploration of different variants. For example, a RefineNet block can accept input from multiple ResNet blocks. We will analyse a 2-cascaded version, a single-block approach as well as a 2-scale 7-path architecture later in~\Sec\ref{sec:alt-arch}.

We denote RefineNet-$m$ as the RefineNet block that connects to the output of block-$m$ in ResNet.
In practice, each ResNet output is passed through one convolutional layer to adapt the dimensionality.
Although all RefineNets share the same internal architecture, their parameters are not tied, allowing for a more flexible adaptation for individual levels of detail.
Following the illustration in \Fig~\refOurFig bottom up, we start from the last block in ResNet, and connect the output of ResNet block-4 to RefineNet-4. Here, there is only one input for RefineNet-4, and RefineNet-4 serves as an extra set of convolutions which adapt the pre-trained ResNet weights to the task at hand, in our case, semantic segmentation.
In the next stage, the output of RefineNet-4 and the ResNet block-3 are fed to RefineNet-3 as 2-path inputs.
The goal of RefineNet-3 is to use the high-resolution features from ResNet block-3 to refine the low-resolution feature map output by RefineNet-4 in the previous stage.
Similarly, RefineNet-2 and RefineNet-1 repeat this stage-wise refinement by fusing high-level information from the later layers and high-resolution but low-level features from the earlier ones. As the last step, the final high-resolution feature maps are fed to a dense soft-max layer to make the final prediction in the form of a dense score map. This score map is then up-sampled to match the original image using bilinear interpolation.

The entire network can be efficiently trained end-to-end. It is important to note that we introduce long-range residual connections between the blocks in ResNet and the RefineNet modules.
During the forward pass, these long-range residual connections convey the low-level features that encode visual details for refining the coarse high-level feature maps.
In the training step, the long-range residual connections allow direct gradient propagation to early convolution layers, which helps effective end-to-end training.

\subsection{RefineNet}
\label{sec:refinet_details}

The architecture of one RefineNet block is illustrated in Fig.~\ref{fig:refine-net-detailed-joint}(a).
In the multi-path overview shown in Fig~\refOurFig,  
RefineNet-1 has one input path, while all other RefineNet blocks have two inputs. Note, however, that our architecture is generic and each Refine block can be easily modified to accept an arbitrary number of feature maps with arbitrary resolutions and depths.

\myparagraph{Residual convolution unit.}
The first part of each RefineNet block consists of an adaptive convolution set that  mainly fine-tunes the pretrained ResNet weights for our task. To that end, each input path is passed sequentially through two residual convolution units (RCU), which is a simplified version of the convolution unit in the original ResNet~\cite{He:2016:ResNet}, where the batch-normalization layers are removed (\cf~\Fig~\ref{fig:refine-net-detailed-joint}(b)). The filter number for each input path is set to 512  for \mbox{RefineNet-4} and 256 for the remaining ones in our experiments.

\myparagraph{Multi-resolution fusion.}
All path inputs are then fused into a high-resolution feature map by the multi-resolution fusion block, depicted in \Fig~\ref{fig:refine-net-detailed-joint}(c).
This block first applies convolutions for input adaptation, which generate feature maps of the same feature dimension (the smallest one among the inputs),
and then up-samples all (smaller) feature maps to the largest resolution of the inputs. Finally, all features maps are fused by summation.
The input adaptation in this block also helps to re-scale the feature values appropriately along different paths, which is important for the subsequent sum-fusion.
If there is only one input path (e.g., the case of RefineNet-4 in \Fig~\ref{fig:refine-net}(c)), the input path will directly go through this block without changes.

\myparagraph{Chained residual pooling.}
The output feature map then goes through the chained residual pooling block, schematically depicted in \Fig~\ref{fig:refine-net-detailed-joint}(d).
The proposed chained residual pooling aims to capture background context from a large image region.
It is able to efficiently pool features with multiple window sizes and fuse them together using learnable weights.
In particular, this component is built as a chain of multiple pooling blocks, each consisting of one max-pooling layer and one convolution layer.
One pooling block takes the output of the previous pooling block as input. 
Therefore, the current pooling block is able to re-use the result from the previous pooling operation and thus access the features from a large region without using a large pooling window.
If not further specified, we use two pooling blocks each with stride 1 in our experiments.

The output feature maps of all pooling blocks are fused together with the input feature map through summation of residual connections.
Note that, our choice to employ residual connections also persists in this building block, which once again facilitates gradient propagation during training.
In one pooling block, each pooling operation is followed by convolutions which serve as a weighting layer for the summation fusion.
It is expected that this convolution layer will learn to accommodate the importance of the pooling block during the training process.

\myparagraph{Output convolutions.}
The final step of each RefineNet block is another residual convolution unit (RCU). This results in a sequence of three RCUs between each block. To reflect this behavior in the last RefineNet-1 block, we place two additional RCUs before the final softmax prediction step.
The goal here is to employ non-linearity operations on the multi-path fused feature maps to generate features for further processing or for final prediction.
The feature dimension remains the same after going through this block.

\subsection{Identity Mappings in RefineNet}
Note that all convolutional components of the RefineNet have been carefully constructed inspired by the idea behind residual connections and follow the rule of identity mapping~\cite{he2016identity}. 
This enables effective backward propagation of the gradient through RefineNet and facilitates end-to-end learning of cascaded multi-path refinement networks.

Employing residual connections with identity mappings allows the gradient to be directly propagated from one block to any other blocks, as was recently shown by~\cite{he2016identity}.  This concept encourages to maintain a clean information path for shortcut connections,
so that these connections are not ``blocked'' by any non-linear layers or components. Instead, non-linear operations are placed on branches of the main information path.
We follow this guideline for developing the individual components in RefineNet, including all convolution units.
It is this particular strategy that allows the multi-cascaded RefineNet to be trained effectively.
Note that we include one non-linear activation layer (ReLU) in the chained residual pooling block. We observed that this ReLU is important for the effectiveness of subsequent pooling operations and it also makes the model less sensitive to changes in the learning rate. We observed that one single ReLU in each RefineNet block does not noticeably reduce the effectiveness of gradient flow.

We have both short-range and long-range residual connections in RefineNet.
Short-range residual connections refer to local shot-cut connections in one RCU or the residual pooling component,
while long-range residual connections refer to the connection between RefineNet modules and the ResNet blocks.
With long-range residual connections,
the gradient can be directly propagated to early convolution layers in ResNet and thus enables end-to-end training of all network components.

The fusion block fuses the information of multiple shortcut paths, which can be considered as performing summation fusion of multiple residual connections with necessary dimension or resolution adaptation. In this aspect, the role of the multi-resolution fusion block here is analogous to the role of the ``summation" fusion in a conventional residual convolution unit in ResNet.
There are certain layers in RefineNet, and in particular within the fusion block, that perform linear feature transformation operations, like linear feature dimension 
reduction or bilinear up-sampling.
These layers are placed on the shortcut paths,
which is similar to the case in ResNet~\cite{He:2016:ResNet}. 
As in in ResNet, when a shortcut connection crosses two blocks, it will include a convolution layer in the shortcut path for linear feature dimension adaptation, 
which ensures that the feature dimension matches the subsequent summation in the next block. 
Since only linear transformation are employed in these layers, gradients still can be propagated through these layers effectively.

\section{Experiments}
\label{sec:experiments}
To show the effectiveness of our approach, we carry out comprehensive experiments on seven public datasets, which include six popular datasets for semantic segmentation on indoors and outdoors scenes (NYUDv2, PASCAL VOC 2012, SUN-RGBD, PASCAL-Context, Cityscapes, ADE20K MIT), and one dataset for object parsing called Person-Part.
The segmentation quality is measured by the intersection-over-union (IoU) score \cite{everingham2010pascal}, the pixel accuracy and the mean accuracy \cite{LongSD14} over all classes.
As commonly done in the literature, we apply simple data augmentation during training. Specifically, we perform random scaling (ranging from $0.7$ to $1.3$), random cropping and horizontal flipping of the images.
If not further specified, we apply test-time multi-scale evaluation, which is a common practice in segmentation methods~\cite{Dai2015arXiv,ChenPK0Y16}.
For multi-scale evaluation, we average the predictions on the same image across different scales for the final prediction.
We also present an ablation experiment to inspect the impact of various components and an alternative 2-cascaded version of our model.
Our system is built on MatConvNet \cite{matconvnet}.

\subsection{Object Parsing}
\label{sec:object-parsing}

\begin{table}[t]
\caption{Object parsing results on the Person-Part dataset.
Our method achieves the best performance (bold).
}
\centering
\resizebox{.5\linewidth}{!}
  {
  \begin{tabular}{ r |  c }
method  & IoU\\ \hline \hline
Attention \cite{chen2015attention} & 56.4 \\
HAZN \cite{xia2015zoom} & 57.5 \\
LG-LSTM \cite{liang2015semantic} & 58.0 \\
Graph-LSTM \cite{liang2016semantic} & 60.2 \\
DeepLab \cite{ChenPKMY14}  &  62.8 \\
DeepLab-v2 (Res101) \cite{ChenPK0Y16}  & 64.9 \\ \hline \hline
\bf RefineNet-Res101 (ours) & \best 68.6

 \end{tabular}
  }
\label{tab:personpart}
\end{table}

\begin{table}[t]
\caption{Ablation experiments on NYUDv2 and Person-Part.
}
\centering
\resizebox{.95\linewidth}{!}
  {
  \begin{tabular}{ c   c  c | c  c  }
Initialization  & Chained pool.  &Msc Eva & NYUDv2 & Person-Parts \\ \hline \hline
ResNet-50  &no &no &40.4 & 64.1 \\
ResNet-50  &yes  &no &42.5 & 65.7 \\
ResNet-50  &yes  &yes  &43.8 & 67.1\\ \hline
ResNet-101 &yes  &no  &43.6 & 67.6\\
ResNet-101 &yes  &yes  &\secbest 44.7 &\secbest 68.6\\ \hline
ResNet-152 &yes  &yes  & \best 46.5 & \best 68.8\\
 \end{tabular}
  }
\label{tab:nyud_ablation}
\end{table}

We first present our results on the task of object parsing, which consists of recognizing and segmenting object parts.
We carry out experiments on the Person-Part dataset~\cite{chen2014detect,chen2015attention} which provides pixel-level labels for six person parts including
Head, Torso, Upper/Lower Arms and Upper/Lower Legs. The rest of each image is considered background. There are training 1717 images and 1818 test images.
We use four pooling blocks in our chained residual pooling for this dataset.

We compare our results to a number of state-of-the-art methods, listed in Table~\ref{tab:personpart}. The results clearly demonstrate the improvement over previous works.
In particular, we significantly outperform
the the recent DeepLab-v2 approach~\cite{ChenPK0Y16} which is based on dilated convolutions for high-resolution segmentation, using the same ResNet as initialization.
In Table~\ref{tab:nyud_ablation}, we present an ablation experiment to quantify the influence of the following components: Network depth, chained residual pooling and multi-scale evaluation (Msc Eva), as described earlier. 
This experiment shows that each of these three factors can improve the overall performance.
Qualitative examples of our object parsing on this dataset are  shown in Fig.\ref{fig:example_personpart}.

\ifx\imglistflag\undefined
\def\imglistflag{}
\newcounter{imgidx}
\newcounter{cntone}
\newcounter{cnttwo}
\newcounter{img_total_one}
\newcounter{img_total_two}
\newcounter{cntthree}
\newcounter{img_total_three}
\fi

 \setcounter{cntone}{0}
 \setcounter{cnttwo}{0}
 \setcounter{img_total_one}{0}
 \setcounter{img_total_two}{0}

\providecommand\settextone[2]{%
  \csdef{textone#1}{#2}}
\providecommand\addtextone[1]{%
  \stepcounter{cntone}%
  \csdef{textone\thecntone}{#1}}
\providecommand\gettextone[1]{%
  \csuse{textone#1}}

\providecommand\settexttwo[2]{%
  \csdef{texttwo#1}{#2}}
\providecommand\addtexttwo[1]{%
  \stepcounter{cnttwo}%
  \csdef{texttwo\thecnttwo}{#1}}
\providecommand\gettexttwo[1]{%
  \csuse{texttwo#1}}

\addtextone{2008_000740}
\addtextone{2010_002927}
\addtextone{2010_005079}
\addtextone{2010_005626}

\addtextone{2010_003287}
\addtextone{2008_002932}

\setcounter{img_total_one}{\arabic{cntone}}
\stepcounter{img_total_one}

\setcounter{img_total_two}{\arabic{cnttwo}}
\stepcounter{img_total_two}

\begin{figure}[t]
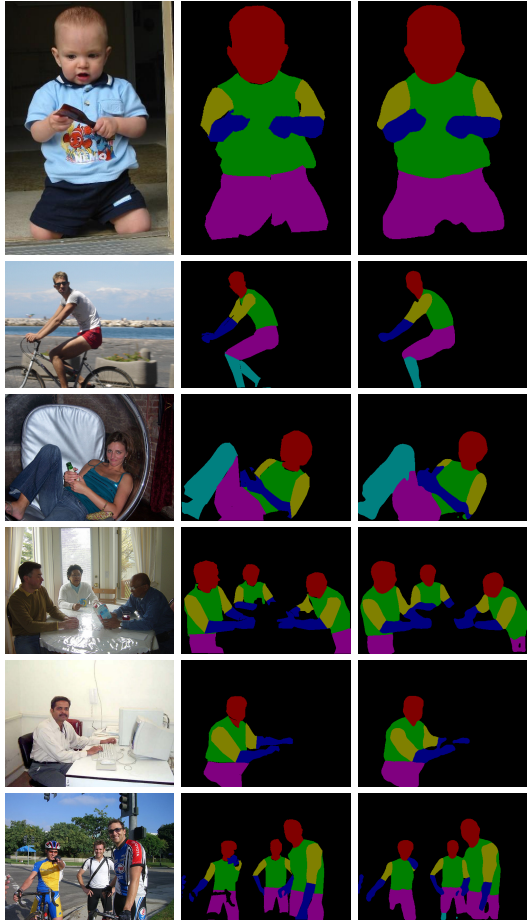

\centering
 \resizebox{.85\linewidth}{!} {
    \begin{subfigure}{1.0in}
	\forloop{imgidx}{1}{\value{imgidx} < \value{img_total_one}}{
		\includegraphics[width=1.0in]{examples/personpart/img/{\gettextone{\arabic{imgidx}}}}\vspace{2pt}
	}
  \vspace{-1.5em}
    \caption{Test Image}
    \end{subfigure}\hspace{1pt}
    \centering
    \begin{subfigure}{1.0in}
	\forloop{imgidx}{1}{\value{imgidx} < \value{img_total_one}}{
		\includegraphics[width=1.0in]{examples/personpart/gt/{\gettextone{\arabic{imgidx}}}}\vspace{2pt}
	}
  \vspace{-1.5em}
    \caption{Ground Truth}
    \end{subfigure}\hspace{1pt}
    \centering
    \begin{subfigure}{1.0in}
    \forloop{imgidx}{1}{\value{imgidx} < \value{img_total_one}}{
		\includegraphics[width=1.0in]{examples/personpart/predict/{\gettextone{\arabic{imgidx}}}}\vspace{2pt}
	}
  \vspace{-1.5em}
    \caption{Prediction}
    \end{subfigure}
 }
 \vspace{0.5em}
    \caption{Our prediction examples on Person-Parts dataset.}
    \label{fig:example_personpart}
\end{figure}

\subsection{Semantic Segmentation}

We now describe our experiments on dense semantic labeling on six public benchmarks and show that our RefineNet outperforms previous methods on all datasets.

\myparagraph{NYUDv2.}
The NYUDv2 dataset~\cite{silberman2012indoor} consists of 1449 RGB-D images showing interior scenes. We use the segmentation labels provided in \cite{gupta2013perceptual}, in which all labels are mapped to $40$ classes.
We use the standard training/test split with $795$ and $654$ images, respectively.
We train our models only on RGB images without using the depth information.
Quantitative results are shown in Table~\ref{tab:nyud}. Our RefineNet achieves new state-of-the-art result on the NYUDv2 dataset.
\begin{table}[t]
\caption{Segmentation results on NYUDv2 (40 classes).
}
\centering
\resizebox{.95\linewidth}{!}
  {
  \begin{tabular}{ r | c | c c c }
method  &training data  &pixel acc. &mean acc. &IoU\\ \hline \hline
Gupta et al. \cite{gupta2014learning}   &RGB-D  &60.3 &-  &28.6\\
FCN-32s \cite{LongSD14} &RGB  &60.0 &42.2 &29.2\\
FCN-HHA \cite{LongSD14} &RGB-D  &65.4 &46.1 &34.0\\ 
Context \cite{lin2016piece} &RGB  &\secbest 70.0  & \secbest 53.6  & \secbest 40.6\\ \hline \hline
\bf RefineNet-Res152 &RGB  & \best 73.6  & \best 58.9  & \best 46.5\\
 \end{tabular}
  }
\label{tab:nyud}
\end{table}

Similar to the object parsing task above, we also perform ablation experiments on the NYUDv2 dataset to evaluate the effect of different settings. The results are presented in Table~\ref{tab:nyud_ablation}.
Once again, this study demonstrates the benefits of adding the proposed chained residual pooling component and deeper networks, both of which consistently improve the performance as measured by IoU.

\myparagraph{PASCAL VOC 2012}
\cite{everingham2010pascal} is a well-known segmentation dataset which includes 20 object categories and one background class.
This dataset is split into a training set, a validation set and a test set,
with $1464$, $1449$ and $1456$ images each. 
Since the test set labels are not publicly available, all reported results have been obtained from the VOC evaluation server.
Following the common convention~\cite{ChenPKMY14,ChenPK0Y16,zheng2015conditional,LiuDPN}, 
the training set is augmented by additional annotated VOC images provided in \cite{HariharanABMM11} as well as with the training data from the MS COCO dataset \cite{lin2014microsoft}.
We compare our RefineNet on the PASCAL VOC 2012 test set with
a number of competitive methods, showing superior performance.
We use dense CRF method in \cite{krahenbuhl2012efficient} for further refinement for this dataset, which gives marginal improvement of $0.1\%$ on the validation set.
Since dense CRF only brings very minor improvement on our high-resolution prediction, we do not apply it on other datasets.

The detailed results for each category and the mean IoU scores are shown in Table \ref{tab:voc12_test_details}.
We achieve an IoU score of $83.4$, which is the best reported result on this challenging dataset to date.\footnote{The result link to the VOC evaluation server: \url{http://host.robots.ox.ac.uk:8080/anonymous/B3XPSK.html}}
We outperform competing methods in almost all categories. 
In particular, we significantly outperform the method DeepLab-v2~\cite{ChenPK0Y16} which is the currently best known dilation convolution method and uses the same ResNet-101 network as initialization.
Selected prediction examples are shown in \Fig~\ref{fig:example_voc}.

\begin{table}[t]
\caption{Segmentation results on the Cityscapes \emph{test} set.
our method achieves the best performance.
}
\centering
\resizebox{.55\linewidth}{!}
  {
  \begin{tabular}{ r |  c }
Method	& IoU\\ \hline \hline
FCN-8s \cite{LongSD14} & 65.3  \\
DPN \cite{LiuDPN} & 66.8 \\
Dilation10 \cite{YuK15}  & 67.1 \\
Context \cite{lin2016piece} &  71.6 \\
LRR-4x \cite{Ghiasi2016} & \secbest 71.8 \\
DeepLab \cite{ChenPKMY14} & 63.1 \\
DeepLab-v2(Res101) \cite{ChenPK0Y16}  & 70.4 \\ \hline \hline
\bf RefineNet-Res101 (ours) & \best 73.6 \\ 
 \end{tabular}
  }
\label{tab:cityscape}
\end{table}

\begin{table*}[tb]
\caption{Results on the PASCAL VOC 2012 test set (IoU scores). Our RefineNet archives the best performance (IoU $83.4$).}
\centering
\resizebox{1\linewidth}{!}
  {
  \begin{tabular}{ r || c c c c c c c c c c c c c c c c c c c c || r }

Method & \rot{aero}  &\rot{bike} &\rot{bird} &\rot{boat} &\rot{bottle}   &\rot{bus}  &\rot{car}  &\rot{cat}  &\rot{chair}    &\rot{cow}  &\rot{table}    &\rot{dog}  &\rot{horse}    &\rot{mbike}    &\rot{person}   &\rot{potted}   &\rot{sheep}    &\rot{sofa} &\rot{train}    &\rot{tv}  & \bf mean \\ \hline \hline
FCN-8s \cite{LongSD14}    &76.8   &34.2   &68.9   &49.4   &60.3   &75.3   &74.7   &77.6   &21.4   &62.5   &46.8   &71.8   &63.9   &76.5   &73.9   &45.2   &72.4   &37.4   &70.9 & 55.1 & 62.2 \\
DeconvNet \cite{noh2015learning} &89.9 &39.3 &79.7 &63.9 &68.2 &87.4 &81.2 &86.1 &28.5 &77.0 &62.0 &79.0 &80.3 &83.6 &80.2 &58.8 & 83.4 &54.3 &80.7 &65.0 &72.5\\
CRF-RNN \cite{zheng2015conditional} &90.4 &55.3 &88.7 &68.4 &69.8 &88.3 &82.4 &85.1 &32.6 &78.5 &64.4 &79.6 &81.9 & 86.4 &81.8 &58.6 &82.4 &53.5 &77.4 &70.1 &74.7\\
BoxSup \cite{Dai2015arXiv} &89.8  &38.0 & 89.2 & 68.9 &68.0 &89.6 &83.0 &87.7 &34.4 &83.6 & 67.1 &81.5 &83.7 &85.2 &83.5 &58.6 &84.9 &55.8 & 81.2 &70.7 &75.2\\
DPN \cite{LiuDPN} &89.0 &  61.6 &87.7 &66.8 &74.7 &91.2 & 84.3 &87.6 &36.5 & 86.3 &66.1 &84.4 &87.8 &85.6 &85.4 &63.6 &87.3 &61.3 &79.4 &66.4 &77.5 \\
Context \cite{lin2016piece} & 94.1 &40.7 &84.1 &67.8 & 75.9 & 93.4 & 84.3 & 88.4 & \secbest 42.5 & 86.4 &64.7 & 85.4 & 89.0 & 85.8  & 86.0 & 67.5 & 90.2 & 63.8 &80.9 & 73.0 & 78.0 \\  
DeepLab \cite{ChenPKMY14} &89.1 &38.3 &88.1 &63.3 &69.7 &87.1 &83.1 &85.0 &29.3 &76.5 &56.5 &79.8 &77.9 &85.8 &82.4 &57.4 &84.3 &54.9 &80.5 &64.1 &72.7 \\
DeepLab2-Res101 \cite{ChenPK0Y16} & 92.6  &60.4 &91.6 &63.4 &76.3 & 95.0 &88.4 &92.6 &32.7 &88.5 &67.6 &89.6 & 92.1 & 87.0 &87.4 &63.3 &88.3 &60.0 & 86.8 &74.5 & 79.7 \\
CSupelec-Res101 \cite{ChandraEccv2016} &92.9 &\secbest 61.2 &91.0 &66.3 &77.7 &\best 95.3 &\best 88.9 &92.4 &33.8 &88.4 &69.1 &89.8 &\best 92.9 &\secbest  87.7 &87.5 &62.6 &89.9 &59.2 &\secbest 87.1 &74.2 &80.2 \\
\hline \hline
\bf RefineNet-Res101   &\best 94.9  &60.2 &\secbest 92.8 & \best 77.5 & \secbest  81.5 &95.0 &87.4 & \secbest  93.3 &39.6 &\secbest 89.3 & \secbest  73.0 &\best 92.7 & \secbest  92.4 &85.4 &\best 88.3 &\secbest 69.7 &\secbest 92.2 &\secbest 65.3 &84.2 &\best 78.7 &\secbest 82.4 \\
\bf RefineNet-Res152   &\secbest 94.7 &\best 64.3 & \best 94.9 & \secbest  74.9 & \best 82.9 &  \secbest  95.1 & \secbest 88.5 & \best 94.7 & \best 45.5 & \best 91.4 & \best 76.3 & \secbest 90.6 &91.8 &\best 88.1 & \secbest 88.0 & \best 69.9 &\best 92.3 & \best 65.9 & \best 88.7 &  \secbest 76.8 & \best 83.4\\

\end{tabular}
  }
\label{tab:voc12_test_details}
\end{table*}

\ifx\imglistflag\undefined
\def\imglistflag{}
\newcounter{imgidx}
\newcounter{cntone}
\newcounter{cnttwo}
\newcounter{img_total_one}
\newcounter{img_total_two}
\newcounter{cntthree}
\newcounter{img_total_three}
\fi

 \setcounter{cntone}{0}
 \setcounter{cnttwo}{0}
 \setcounter{img_total_one}{0}
 \setcounter{img_total_two}{0}

\providecommand\settextone[2]{%
  \csdef{textone#1}{#2}}
\providecommand\addtextone[1]{%
  \stepcounter{cntone}%
  \csdef{textone\thecntone}{#1}}
\providecommand\gettextone[1]{%
  \csuse{textone#1}}

\providecommand\settexttwo[2]{%
  \csdef{texttwo#1}{#2}}
\providecommand\addtexttwo[1]{%
  \stepcounter{cnttwo}%
  \csdef{texttwo\thecnttwo}{#1}}
\providecommand\gettexttwo[1]{%
  \csuse{texttwo#1}}

\addtextone{2009_003071}
\addtextone{2008_003461}
\addtextone{2007_001311}
\addtextone{2010_003854}
\addtextone{2007_006449}

\addtextone{2007_004902}
\addtextone{2009_003576}

\setcounter{img_total_one}{\arabic{cntone}}
\stepcounter{img_total_one}

\setcounter{img_total_two}{\arabic{cnttwo}}
\stepcounter{img_total_two}

\begin{figure}[t]
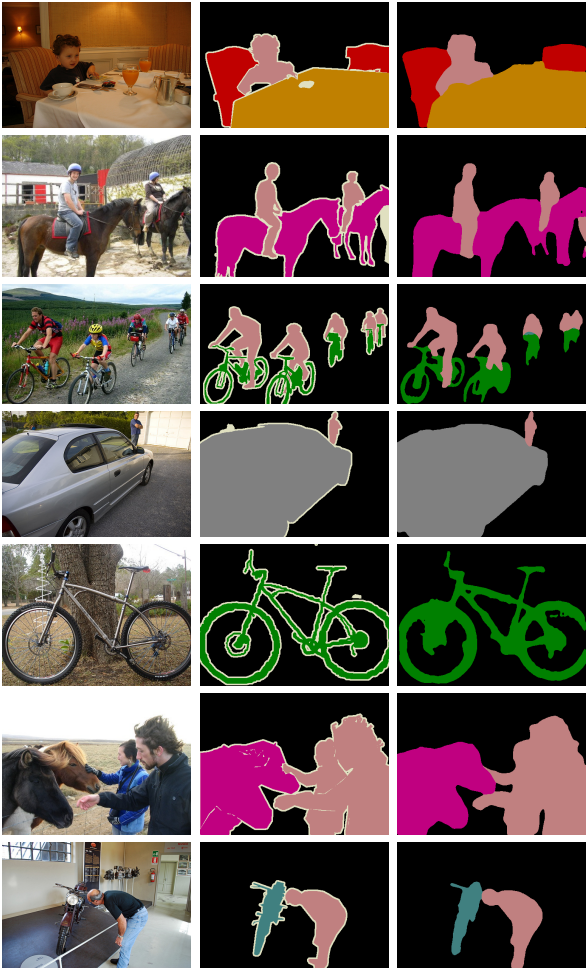

\centering
 \resizebox{.95\linewidth}{!} {
    \begin{subfigure}{1.0in}
	\forloop{imgidx}{1}{\value{imgidx} < \value{img_total_one}}{
		\includegraphics[width=1.0in]{examples/voc/img/{\gettextone{\arabic{imgidx}}}}\vspace{2pt}
	}
  \vspace{-1.5em}
    \caption{Test Image}
    \end{subfigure}\hspace{1pt}
    \centering
    \begin{subfigure}{1.0in}
	\forloop{imgidx}{1}{\value{imgidx} < \value{img_total_one}}{
		\includegraphics[width=1.0in]{examples/voc/gt/{\gettextone{\arabic{imgidx}}}}\vspace{2pt}
	}
  \vspace{-1.5em}
    \caption{Ground Truth}
    \end{subfigure}\hspace{1pt}
    \centering
    \begin{subfigure}{1.0in}
    \forloop{imgidx}{1}{\value{imgidx} < \value{img_total_one}}{
		\includegraphics[width=1.0in]{examples/voc/predict/{\gettextone{\arabic{imgidx}}}}\vspace{2pt}
	}
  \vspace{-1.5em}
    \caption{Prediction}
    \end{subfigure}
 }
 \vspace{0.5em}
    \caption{Our prediction examples on VOC 2012 dataset.}
    \label{fig:example_voc}
\end{figure}

\ifx\imglistflag\undefined
\def\imglistflag{}
\newcounter{imgidx}
\newcounter{cntone}
\newcounter{cnttwo}
\newcounter{img_total_one}
\newcounter{img_total_two}
\newcounter{cntthree}
\newcounter{img_total_three}
\fi

 \setcounter{cntone}{0}
 \setcounter{cnttwo}{0}
 \setcounter{img_total_one}{0}
 \setcounter{img_total_two}{0}

\providecommand\settextone[2]{%
  \csdef{textone#1}{#2}}
\providecommand\addtextone[1]{%
  \stepcounter{cntone}%
  \csdef{textone\thecntone}{#1}}
\providecommand\gettextone[1]{%
  \csuse{textone#1}}

\providecommand\settexttwo[2]{%
  \csdef{texttwo#1}{#2}}
\providecommand\addtexttwo[1]{%
  \stepcounter{cnttwo}%
  \csdef{texttwo\thecnttwo}{#1}}
\providecommand\gettexttwo[1]{%
  \csuse{texttwo#1}}

\addtextone{lindau_000026_000019}
\addtextone{munster_000051_000019}
\addtextone{munster_000098_000019}
\addtextone{munster_000078_000019}
\addtextone{frankfurt_000001_059789}
\addtextone{frankfurt_000001_064651}

\setcounter{img_total_one}{\arabic{cntone}}
\stepcounter{img_total_one}

\setcounter{img_total_two}{\arabic{cnttwo}}
\stepcounter{img_total_two}

\begin{figure}[t]
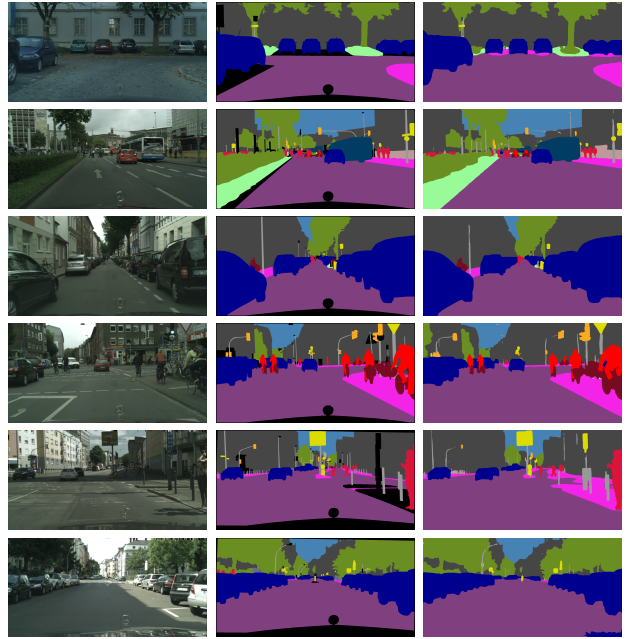

\centering
 \resizebox{1\linewidth}{!} {
    \begin{subfigure}{1.0in}
	\forloop{imgidx}{1}{\value{imgidx} < \value{img_total_one}}{
		\includegraphics[width=1.0in]{examples/city/img/{\gettextone{\arabic{imgidx}}}}\vspace{2pt}
	}
  \vspace{-1.5em}
    \caption{Test Image}
    \end{subfigure}\hspace{1pt}
    \centering
    \begin{subfigure}{1.0in}
	\forloop{imgidx}{1}{\value{imgidx} < \value{img_total_one}}{
		\includegraphics[width=1.0in]{examples/city/gt/{\gettextone{\arabic{imgidx}}}}\vspace{2pt}
	}
  \vspace{-1.5em}
    \caption{Ground Truth}
    \end{subfigure}\hspace{1pt}
    \centering
    \begin{subfigure}{1.0in}
    \forloop{imgidx}{1}{\value{imgidx} < \value{img_total_one}}{
		\includegraphics[width=1.0in]{examples/city/predict/{\gettextone{\arabic{imgidx}}}}\vspace{2pt}
	}
  \vspace{-1.5em}
    \caption{Prediction}
    \end{subfigure}
 }
 \vspace{0.5em}
    \caption{Our prediction examples on Cityscapes dataset.}
    \label{fig:example_city}
\end{figure}

\myparagraph{Cityscapes}
\cite{Cordts2016Cityscapes} is a very recent dataset on street scene images from $50$ different European cities.
This dataset provides fine-grained pixel-level annotations of roads, cars, pedestrians, bicycles, sky, \etc.
The provided training set has $2975$ images and the validation set has $500$ images.
In total, $19$ classes are considered for training and evaluation.
The test set ground-truth is withheld by the organizers, and we evaluate our method on the their evaluation server.
The test results are shown in Table~\ref{tab:cityscape}.
In this challenging setting, our architecture again outperforms previous methods.
A few test images along with ground truth and our predicted semantic maps are shown in Fig.~\ref{fig:example_city}.

\myparagraph{PASCAL-Context.}
The PASCAL-Context \cite{mottaghi2014role} dataset provides the segmentation labels of the whole scene for the PASCAL VOC images.
We use the segmentation labels which contain $60$ classes ($59$ object categories plus background) for evaluation as well as the provided training/test splits. The training set contains $4998$ images and the test set has $5105$ images.
Results are shown in Table \ref{tab:pascalcontext}. Even without additional training data and with the same underlying ResNet architecture with 101 layers, we outperform the previous state-of-the-art achieved by DeepLab.

\begin{table}[t]
\caption{Segmentation results on PASCAL-Context dataset (60 classes).
Our method performs the best. We only use the VOC training images. }
\centering
\resizebox{.75\linewidth}{!}
  {
  \begin{tabular}{ r | c c c c} 
  Method	& Extra train data 	&IoU\\ \hline \hline
O2P \cite{carreira2012semantic} &- 	&18.1\\
CFM \cite{dai2014convolutional} &-	&34.4\\
FCN-8s \cite{LongSD14} &- 	&35.1\\
BoxSup \cite{Dai2015arXiv} &-	&40.5\\ 
HO-CRF \cite{arnab2016higher} &- &41.3\\ 
Context \cite{lin2016piece} &-	& 43.3\\ 
DeepLab-v2(Res101) \cite{ChenPK0Y16} & COCO ($\sim$100K) & 45.7 \\ \hline \hline
RefineNet-Res101 (ours) &- & \secbest 47.1 \\
RefineNet-Res152 (ours) &- & \best 47.3\\
 \end{tabular}
  }
\label{tab:pascalcontext}
\end{table}

\myparagraph{SUN-RGBD}
\cite{song2015sun} is a segmentation dataset that contains around $10,000$ RGB-D indoor images and provides pixel labeling masks for $37$ classes.
Results are shown in Table \ref{tab:sunrgbd}. Our method outperforms
all existing methods by a large margin across all evaluation metrics, even though we do not make use of the depth information for training.

\begin{table}[t]
\caption{Segmentation results on SUN-RGBD dataset (37 classes).
We compare to a number of recent methods.
Our RefineNet significantly outperforms the existing methods.}
\centering
\resizebox{.95\linewidth}{!}
  {
  \begin{tabular}{ r | c | c c c }
  Method  & Train data  & Pixel acc. & Mean acc. &IoU\\ \hline \hline
  Liu et al. \cite{liu2011sift}  &RGB-D  & $-$ &10.0 & $-$ \\
  Ren et al. \cite{ren2012rgb}  &RGB-D  & $-$ &36.3 & $-$ \\
  Kendall et al. \cite{KendallBC15} &RGB  &71.2 &45.9 &30.7 \\ 
  Context \cite{lin2016piece}  &RGB  & 78.4 & 53.4 & 42.3 \\ \hline \hline
RefineNet-Res101 &RGB  & \secbest 80.4 & \secbest 57.8 & \secbest 45.7 \\
RefineNet-Res152  & RGB & \best 80.6 & \best 58.5 & \best 45.9 \\
 \end{tabular}
  }
\label{tab:sunrgbd}
\end{table}

\myparagraph{ADE20K MIT}
\cite{ZhouZPFBT16} is a newly released dataset for scene parsing which provides dense labels of 150 classes on more than 20K scene images. The categories include a large variety of objects (\eg, person, car, \etc) and stuff (\eg, sky, road, \etc). The provided validation set consisting of 2000 images is used for quantitative evaluation. Results are shown in Table \ref{tab:ADE}. 
Our method clearly outperforms the baseline methods described in \cite{ZhouZPFBT16}.

\begin{table}[t]
\caption{Segmentation results on the ADE20K dataset (150 classes)  \emph{val} set.
our method achieves the best performance.
}
\centering
\resizebox{.55\linewidth}{!}
  {
  \begin{tabular}{ r |  c }
Method  & IoU\\ \hline \hline
FCN-8s \cite{LongSD14}   &  29.4\\
SegNet \cite{BadrinarayananK15}  &  21.6\\
DilatedNet \cite{ChenPKMY14,YuK15}   &  32.3\\
Cascaded-SegNet \cite{ZhouZPFBT16}   &  27.5\\
Cascaded-DilatedNet \cite{ZhouZPFBT16}   &  34.9\\ \hline \hline
RefineNet-Res101 (ours) & \secbest 40.2 \\
RefineNet-Res152 (ours) & \best 40.7 \\
 \end{tabular}
  }
\label{tab:ADE}
\end{table}

\begin{figure*}[t]
  \centering  
  \includegraphics[width=1\linewidth]{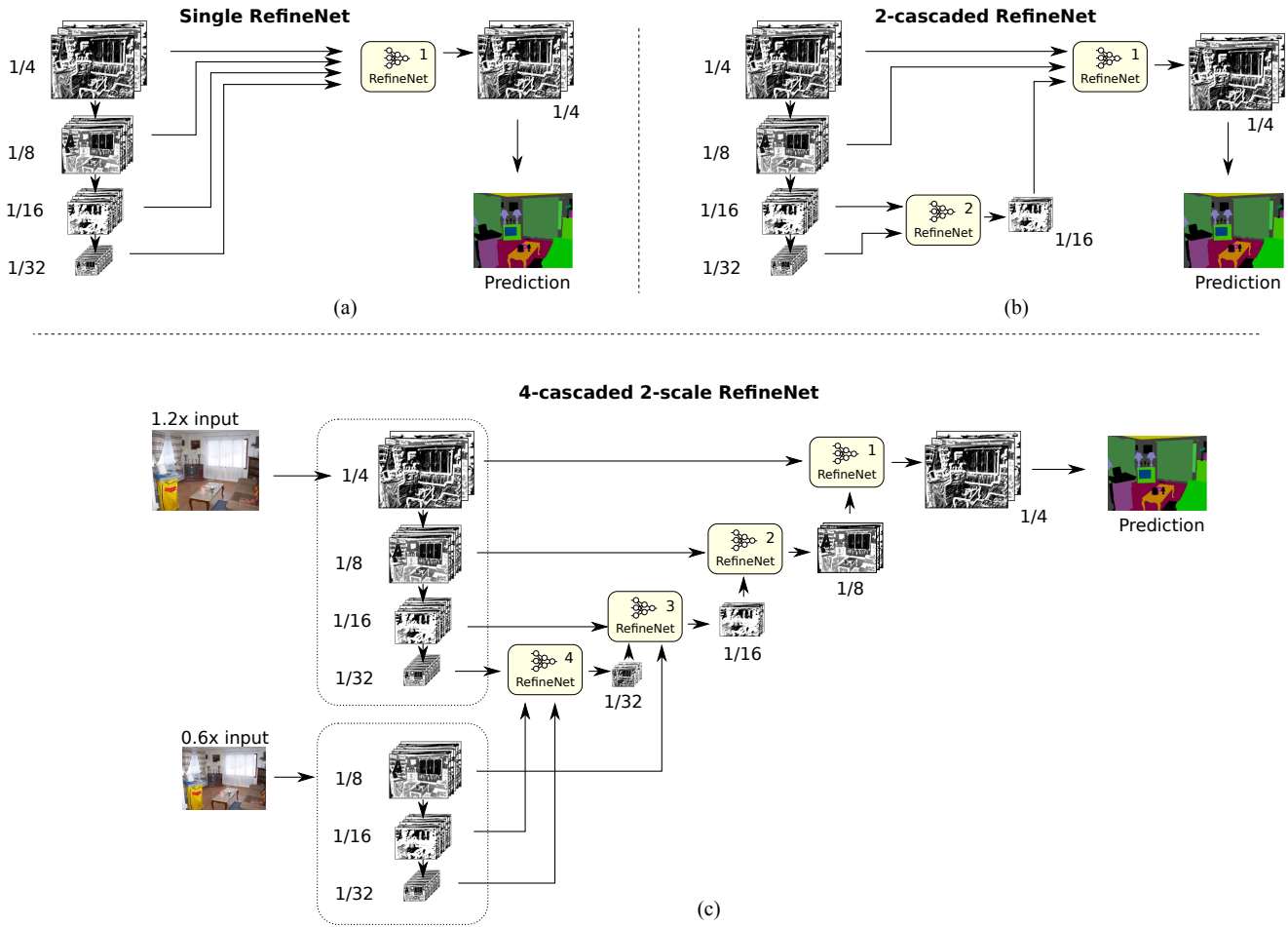}
\caption{Illustration of 3 variants of our network architecture:  (a) single RefineNet, (b) 2-cascaded RefineNet and (c) 4-cascaded RefineNet with 2-scale ResNet. Note that our proposed RefineNet block can seamlessly handle different numbers of inputs of arbitrary resolutions and dimensions without any modification.}
\label{fig:refine-net-variants}
\end{figure*}

\subsection{Variants of cascaded RefineNet}
\label{sec:alt-arch}

\begin{table}
\caption{Evaluations of 4 variants of cascaded RefineNet: single RefineNet, 2-cascaded RefineNet, 4-cascaded RefineNet, 4-cascaded RefineNet with 2-scale ResNet on the NYUDv2 dataset. 
We use the 4-cascaded version as our main architecture throughout all experiments in the paper because this turns out to be the best compromise between accuracy and efficiency.
}
\centering
\resizebox{1\linewidth}{!} 
  {
  \begin{tabular}{ r |  c | c | c }
Variant & Initialization   &Msc Eva &IoU \\ \hline \hline
single RefineNet & ResNet-50  &no & 40.3\\
2-cascaded RefineNet & ResNet-50  &no & 40.9\\
4-cascaded RefineNet & ResNet-50  &no & 42.5\\
4-cascaded 2-scale RefineNet & ResNet-50  &no & 43.1\\
 \end{tabular}
  }
\label{tab:variant}
\end{table}

As discussed earlier, our RefineNet is flexible in that it can be cascaded in various manners for generating various architectures.
Here, we discuss several variants of our RefineNet. Specifically, we present the architectures of using
a single RefineNet, a 2-cascaded RefineNet and a 4-cascaded RefineNet with 2-scale ResNet.
The architectures of all three variants are illustrated in Fig.~\ref{fig:refine-net-variants}.
The architecture of 4-cascaded RefineNet is already presented in \Fig~\refOurFig. 
Please note that this 4-cascaded RefineNet model is the one used in all other experiments.

The single RefineNet model is the simplest variant of our network. It consists of only one single RefineNet block, which takes all four inputs from the four blocks of ResNet and fuses all-resolution feature maps in a single process.  
The 2-cascaded version is similar our main model (4-cascaded) from \Fig~\refOurFig, but employs only two RefineNet modules instead of four. The bottom one, RefineNet-2, has two inputs from ResNet blocks 3 and 4, and the other one has three inputs, two coming from the remaining ResNet blocks and one from RefineNet-2.
For the 2-scale model in Fig.~\ref{fig:refine-net-variants}(c), we use 2 scales of the image as input and respectively 2 ResNets to generate feature maps;
the input image is scaled to a factor of 1.2 and 0.6 and fed into 2 independent ResNets.

The evaluation results of these variants on the NYUD dataset are shown in Table \ref{tab:variant}. 
This experiment demonstrates that the 4-cascaded version yields better performance than the 2-cascaded and 1-cascaded version, and using 2-scale image input with 2 ResNet is better than using 1-scale input. This is expected due to the larger capacity of the network. However, it also results in longer training times. Hence, we resort to using the single-scale 4-cascaded version as the standard architecture in all our experiments.

\section{Conclusion}
\label{sec:conclusion}

We have presented RefineNet, a novel multi-path refinement network  for semantic segmentation and object parsing. The cascaded architecture is able to effectively combine high-level semantics and low-level features to produce high-resolution segmentation maps. 
Our design choices are inspired by the idea of identity mapping which facilitates gradient propagation across long-range connections and thus enables effective end-to-end learning.
We outperform all previous works on seven public benchmarks, setting a new mark for the state of the art in semantic labeling.

\paragraph{Acknowledgments}
This research was supported by the Australian Research Council
through the Australian Centre for Robotic Vision (CE140100016).
C. Shen's participation was supported by an ARC Future Fellowship (FT120100969).
I. Reid's participation was supported by an ARC Laureate Fellowship (FL130100102).

{\small
\bibliographystyle{ieee}
\bibliography{mybib,short,references}
}

\end{document}